\documentclass[runningheads]{llncs}
\usepackage[T1]{fontenc}
\usepackage[tableposition=top]{caption}
\usepackage{subcaption}
\usepackage{graphicx}
\usepackage{amssymb}
\usepackage{amsfonts,mathtools}
\usepackage[ruled]{algorithm2e}
\usepackage{float}
\usepackage{multirow}
\usepackage{color}
\usepackage[table]{xcolor}
\usepackage{soul}
\usepackage{subcaption}
\begin{document}
\title{Structured Radial Basis Function Network: Modelling Diversity for Multiple Hypotheses Prediction}
\titlerunning{Structured Radial Basis Function Network}
%
\author{Alejandro Rodriguez Dominguez \and
Muhammad Shahzad \and Xia Hong}
\authorrunning{A. Rodriguez Dominguez et al.}
%
\institute{Department of Computer Science, University of Reading, United Kingdom
\email{a.j.rodriguezdominguez@pgr.reading.ac.uk \\ m.shahzad2@reading.ac.uk \\ xia.hong@reading.ac.uk}}
%
\maketitle              
\begin{abstract}
Multi-modal problems can be effectively addressed using multiple hypothesis frameworks, but integrating these frameworks into learning models poses significant challenges. This paper introduces a Structured Radial Basis Function Network (s-RBFN) as an ensemble of multiple hypothesis predictors for regression. During the training of the predictors, first the centroidal Voronoi tessellations are formed based on their losses and the true labels, representing geometrically the set of multiple hypotheses. Then, the trained predictors are used to compute a structured dataset with their predictions, including centers and scales for the basis functions. A radial basis function network, with each basis function focused on a particular hypothesis, is subsequently trained using this structured dataset for multiple hypotheses prediction. The s-RBFN is designed to train efficiently while controlling diversity in ensemble learning parametrically. The least-squares approach for training the structured ensemble model provides a closed-form solution for multiple hypotheses and structured predictions. During the formation of the structured dataset, a parameter is employed to avoid mode collapse by controlling tessellation shapes. This parameter provides a mechanism to balance diversity and generalization performance for the s-RBFN. The empirical validation on two multivariate prediction datasets—air quality and energy appliance predictions—demonstrates the superior generalization performance and computational efficiency of the structured ensemble model compared to other models and their single-hypothesis counterparts.
\end{abstract}
\section{Introduction}
\label{Introduction}
Multi-modality focuses on perception with a set of hypotheses instead of a single output to learn processes. Notable existing approaches include Multiple Choice Learning (MCL) \cite{NIPS2012_cfbce4c1,pmlr-v33-guzman-rivera14}, Multiple Hypotheses Prediction (MHP) \cite{rupprecht2017learning}, Mixture-Of-Experts \cite{6215056}, Bagging \cite{10.1023/A:1018054314350}, Boosting \cite{inproceedingsBoosting}, and Meta-Learning \cite{10.1007/s10462-022-10283-5}. Among them, MCL differs in that it uses the output of different models or hypotheses as inputs to a structured ensemble model (or multiple structured prediction/classification task models), which are heterogeneous ensemble predictors that can vary in size, parameters, and architecture \cite{NIPS2012_cfbce4c1}. Diverse Multiple Choice Learning (DivMCL), is an extension proposed for diverse multi-output structured prediction by including a diversity encouraging term in the loss function used for training the models \cite{pmlr-v33-guzman-rivera14}. While DivMCL provides diversity, it trains separate networks which makes information exchange between individual predictors harder. To cope with this, the DivMCL ideas are extended by instead of training separate networks for each choice, the individual hypotheses are combined with Voronoi tessellations formed by the predictors’ losses in a shared architecture. This allows sharing of information among predictors during training \cite{rupprecht2017learning}. But, it is not clear how to optimally combine these predictors into an ensemble. To the best of the authors’ knowledge, there is no existing method that optimally combines structured predictions from multiple hypotheses prediction with an ensemble learning model that can be trained with a closed-form solution \cite{NIPS2012_cfbce4c1,pmlr-v33-guzman-rivera14,rupprecht2017learning}.

Another important aspect in enhancing generalization of ensembles is the diversity of individual predictors. Diversity in this context has been extensively researched in literature, e.g., using Bias-Variance-Covariance decomposition \cite{Ueda1996GeneralizationEO}, ambiguity decomposition \cite{NIPS1994_b8c37e33}, and their hybrid extensions \cite{5ab71015f97f4fc48a615bd5e8b673e4}. However, there is also not a unifying framework for diversity in ensemble learning. Moreover, there is no clear connection in the literature between geometric properties of loss functions for individual predictors and diversity in ensemble learning \cite{wood2023unified}. This work focuses on the definition of diversity in ensemble learning as the variety of outputs from base learners that can improve the generalization performance of ensemble models \cite{wood2023unified}.

Building on previous aspects \cite{pmlr-v33-guzman-rivera14,rupprecht2017learning,wood2023unified}, a new approach for multiple hypotheses prediction using a structured ensemble model is presented. In this approach, predictions from a set of base learners or individual predictors are used as inputs for a radial basis function network, with each predictor or hypothesis focusing on a specific basis function. The model is referred to as the Structured Radial Basis Function Network (s-RBFN). During training, the base learners form centroidal Voronoi tessellations (CVT), with each hypothesis or base learner assigned to a particular tessellation. A parametric formula from multiple hypotheses framework \cite{rupprecht2017learning} is used to weight the updates of the base learner parameters in each iteration of gradient descent, preventing mode collapse and ensuring that all predictions fall within their respective tessellations. In this work, this strategy is applied to control diversity in ensemble learning similar to DivMCL \cite{pmlr-v33-guzman-rivera14} with the mechanism from MHP \cite{rupprecht2017learning}, enhancing generalization performance. The proposed s-RBFN can then be optimized using least squares, providing faster training compared to other existing structured models that rely on gradient descent or non-convex methods \cite{NIPS2012_cfbce4c1,pmlr-v33-guzman-rivera14,rupprecht2017learning}.

The paper is organized as follows: Section \ref{litrev} presents a revision of the previous work in structured ensemble learning and diversity; Section \ref{modprop} presents the proposed model; Section \ref{secexp} presents the experimental results and discussions; and finally, Section \ref{conc} provides the concluding remarks and outlook.

\section{Literature Review}
\label{litrev}

Multiple hypotheses prediction (MHP) methods extend semi-supervised ensembles and other single-loss, single-output systems to multiple outputs providing a piece-wise constant approximation of the conditional output space. They differ from mixture density networks by representing the uncertainty through a discrete set of hypotheses \cite{rupprecht2017learning}. These models initially employed training techniques from multiple choice learning \cite{10.1007/978-3-642-33715-4_1,pmlr-v33-guzman-rivera14} and later exploited the geometric properties from Voronoi tessellations formed by losses of the individual predictors as multiple hypothesis \cite{rupprecht2017learning}. These approaches tend to be based on Winner-Takes-it-All (WTA) loss, meaning that the best base learner among all predictors gets updated during their training. A partial solution is a relaxed version of WTA \cite{rupprecht2017learning} where in addition to the winner predictor, the other predictors also get updated for each iteration. It alleviates the convergence problem of the WTA, but still leads to hypotheses with incorrect modes. Moreover, when optimizing for a mixture distribution, the issues of numerical instabilities and mode collapsing arise. For this purpose, the evolving WTA loss was proposed \cite{DBLP:journals/corr/abs-2110-02858} which addresses these issues by preserving the distribution, yielding regularly distributed hypotheses. Although this somewhat mitigates the issue but still the problem of how to combine the multiple hypothesis efficiently in a structured ensemble model persists.

Another aspect of MHP is the use of diversity which can serve as effective regularization - leading to possibly worse performance on training data, but better generalization on unseen test data \cite{pmlr-v33-guzman-rivera14}. Traditional diversity measures often assess the correlation or discrepancy between predictions of two models and their collective performance \cite{articleKuncheva}. Recent innovations have introduced the Bias-Variance-Diversity decomposition, a nuanced framework that integrates various functional forms for each loss and directly links diversity to the expectation of ensemble ambiguity \cite{wood2023unified}. This approach goes beyond the traditional Ambiguity and Bias-Variance-Covariance decompositions, limited to squared-loss and arithmetic-mean combiners \cite{NIPS1994_b8c37e33,Ueda1996GeneralizationEO}. In practice, strategies like bagging and boosting facilitate diversity among base learners by manipulating data, thus introducing structural and data diversity. Additional methods quantify diversity through non-maximal predictions and employ metrics such as the logarithm of ensemble diversity (LED) and ensemble entropy \cite{pmlr-v97-pang19a,10.1007/s10462-022-10283-5}. 
More recently, MCL and DivMCL demonstrate superior test accuracy and better generalization compared to traditional multi-output prediction methods \cite{NIPS2012_cfbce4c1,pmlr-v33-guzman-rivera14}. These approaches emphasize minimizing oracle loss by focusing on specific hypothesis, contrasting with broader Mixture-of-Expert models \cite{pmlr-v97-pang19a,10.1007/s10462-022-10283-5,pmlr-v33-guzman-rivera14,NIPS2012_cfbce4c1,10.5555/3157096.3157334}. Ultimately, the strategic integration of diversity not only serves as an effective regularization mechanism but also critically enhances the predictive accuracy and reliability of ensemble models, especially in managing out-of-distribution data. By optimizing ensemble diversity through sophisticated decomposition models and diverse ensemble strategies, researchers can effectively balance error components to minimize overall mean-squared error, resulting in significantly improved predictions \cite{Ueda1996GeneralizationEO,articlediv}.

\section{Proposed Methodology}
\label{modprop}
In this section, first the multiple hypotheses prediction with Voronoi Tessellations is presented. Later, it is explained how this could be scaled to operate in a structured setting for regression applications. This is done by generating the structured dataset using the MHP base learners' predictions. Finally the optimization of the s-RBFN using the structured dataset is efficiently carried out by least squares approach.

In the supervised learning setting, given training instances \(\{\boldsymbol{x}_i\}_{i=1}^N\) and ground-truth labels \(\{y_i\}_{i=1}^N\), the multiple hypotheses case involves a set of prediction functions \(\{f_{\boldsymbol{\theta}_j}(\boldsymbol{x})\}_{j=1}^M\) with corresponding model parameters \(\boldsymbol{\Theta} = \{\boldsymbol{\theta}_j\}_{j=1}^M\). Assuming the training samples follow the distribution \(p(\boldsymbol{x}, y)\), the expected error for a loss function \(\mathcal{L}\) is expressed as:

\begin{equation}
    \int_{\boldsymbol{X}} \sum_{j=1}^{M} \int_{\mathcal{Y}_j(f_{\boldsymbol{\theta}_j}(\boldsymbol{x}))} \mathcal{L}(f_{\boldsymbol{\theta}_j}(\boldsymbol{x}), y) \, p(\boldsymbol{x}, y) \, \mathrm{d}y \, \mathrm{d}\boldsymbol{x}
    \label{eq14}
\end{equation}

During training, the Voronoi tessellation of the label space is induced by the losses computed from $M$ predictors and given as $\mathcal{Y}=\bigcup_{j=1}^{M}\mathcal{Y}_j(f_{\boldsymbol{\theta}_j}(\boldsymbol{x}))$ where $\mathcal{Y}_j(f_{\boldsymbol{\theta}_j}(\boldsymbol{x_i}))$ represents the \textit{j}th cell with $f_{\boldsymbol{\theta}_j}(\boldsymbol{x_i})$ being the closest of the $M$ predictions to the label data for each training iteration \cite{rupprecht2017learning}:

\begin{equation}
    \mathcal{Y}_j(f_{\boldsymbol{\theta}_j}(\boldsymbol{x_i}))=\left\{y_i\in\mathcal{Y}_j:\mathcal{L}(f_{\boldsymbol{\theta}_j}(\boldsymbol{x_i}),y_i)<\mathcal{L}(f_{\boldsymbol{\theta}_k}(\boldsymbol{x_i}),y_i)\forall k\neq j\right\}
\label{Eq15}
\end{equation}

While implementing (\ref{Eq15}), a typical approach adopted to avoid mode collapse is to relax the best-of-\textit{M} approach by updating all predictors in each iteration \cite{NIPS2012_cfbce4c1,pmlr-v33-guzman-rivera14}. Existing works either focus on multi-output prediction or does not provide an efficient way to combine the base learners or multiple hypotheses, often relying on numerical methods \cite{NIPS2012_cfbce4c1,pmlr-v33-guzman-rivera14,9893798}. To this end, the aim of this work is to efficiently combine, in a structured model, the set of hypotheses that form the centroidal Voronoi tessellations. 
Additionally, the hypothesis that manipulating the shape of the tessellations formed during the training of the predictors, that has direct implications in generalization performance, is validated in the experiments. This is due to the diversity in ensemble learning induced by the predictors.

\subsection{Structured Dataset Formation}

Two step approach have been taken for structured dataset formation. Firstly, the set of predictors $\{f_{\boldsymbol{\theta}_j}(\boldsymbol{x})\}_{j=1}^M$ are trained with stochastic gradient descent with randomly initialised weights. Secondly, these learned models are used to generate the predictions that form the structured dataset.

In each \textit{i}th iteration, using the \textit{j}th prediction \(f_{\boldsymbol{\theta}_j}(\boldsymbol{x_i})\) and the true label \( y_i \), the predictors' parameters are updated using the stochastic gradient descent as follows: 

\begin{equation}
{\boldsymbol{\theta}}_j = {\boldsymbol{\theta}}_j - \eta_j\left(\frac{\partial {\mathcal{L}(f_{\boldsymbol{\theta}_j}(\boldsymbol{x_i}), y_i)}}{\partial {\boldsymbol{\theta}}_j}+\frac{\lambda_p}{N}{\boldsymbol{\theta}}_j\right)\delta\left(\mathcal{Y}_j\left(f_{\boldsymbol{\theta}_j}(\boldsymbol{x_i})\right)\right)
\label{predparam}
\end{equation}

where $\eta_j$ denotes the learning rate for the \textit{j}th predictor and the norm loss is computed as $\mathcal{L}(f_{\boldsymbol{\theta}_j}(\boldsymbol{x_i}), y_i)=\|f_{\boldsymbol{\theta}_j}(\boldsymbol{x_i}) - y_i\|_2^2 + \frac{\lambda_p}{2N} \sum_{j=1}^{M} \boldsymbol{\theta}_j^2$ with the regularization parameter $\lambda_p$. The function $\delta\left(\mathcal{Y}j\left(f{\boldsymbol{\theta}_j}(\boldsymbol{x_i})\right)\right)$ serves as an indicator with a parameter $0<\varepsilon<1$ that can alter the shape of the tessellation during training \cite{rupprecht2017learning}. This parameter enhances diversity in the structured dataset for ensemble generalization by regulating the extent to which non-top predictors' parameters are updated in each training iteration. It is defined as:

\begin{equation}
\delta(y\in\mathcal{Y}_j(f_{\boldsymbol{\theta}_j}(\boldsymbol{x})))=\left\{\begin{matrix}1-\varepsilon& \ 
 \text{if is  true}\\\frac{\varepsilon}{M-1}& \text{otherwise}\\\end{matrix}\right.
\label{equatruedelta}
\end{equation}

When the training of the predictors is completed, the same set of training instances $\{\boldsymbol{x}_i\}_{i=1}^N$ are used to generate structured dataset. To elaborate, if the prediction obtained after the forward pass for \textit{j}th predictor on a particular training instance $\boldsymbol{x}_i$ is denoted as  $f_{\boldsymbol{\theta}_j}(\boldsymbol{x}_i)$, then the resulting predictions for the entire structured dataset can be written in the matrix form as:

\begin{equation}
\boldsymbol{D}(\varepsilon) = \left[\begin{matrix}f_{\boldsymbol{\theta}_1}\left(\boldsymbol{x}_1 \right)&\ldots&f_{\boldsymbol{\theta}_M}\left(\boldsymbol{x}_1 \right)\\\vdots&\ddots&\vdots\\f_{\boldsymbol{\theta}_1}\left(\boldsymbol{x}_N \right)
&\ldots&f_{\boldsymbol{\theta}_M}\left(\boldsymbol{x}_N \right)\\\end{matrix}\right]
\end{equation}

with \(\boldsymbol{D}(\varepsilon) \in \mathbb{R}^{N \times M}\) being the matrix of predictions for a particular diversity parameter $0\leq\varepsilon\leq 1$. 
Similarly, for any test set with test instances $\{\boldsymbol{x}_i^{\prime}\}_{i=1}^n$, the structured test set $\boldsymbol{D}(\varepsilon)^{\prime}\in \mathbb{R}^{n \times M}$ is given by the predictions $\{f_{\boldsymbol{\theta}_j}(\boldsymbol{x}_i^{\prime})\}_{i=1}^n$. For any structured test dataset, the predictors use the same set of parameters $\boldsymbol{\Theta}$ obtained after training using stochastic gradient descent.

\subsection{s-RBFN Optimisation}
The structured dataset is used as input for the radial basis function network, with each \textit{j}th predictor or hypothesis $f_{\boldsymbol{\theta}_j}(\boldsymbol{x})$ associated to a particular basis function $\phi\left(f_{\boldsymbol{\theta}_j}\left(\boldsymbol{x}\right),\mu_j,\sigma_j\right)$, i.e., a map \(\boldsymbol{\Phi}\left(\boldsymbol{D(\varepsilon)}\right): \mathbb{R}^{N \times M} \to \mathbb{R}^{N \times M}\) is obtained by applying the basis function $\phi\left(\cdot\right)$ to each element of \(\boldsymbol{D}(\varepsilon)\), transforming it into:

\[
\boldsymbol{\Phi}\left(\boldsymbol{D(\varepsilon)}\right) = \left[\begin{matrix}
\phi(f_{\boldsymbol{\theta}_1}(\boldsymbol{x}_1), \mu_1, \sigma_1) & \ldots & \phi(f_{\boldsymbol{\theta}_M}(\boldsymbol{x}_1), \mu_M, \sigma_M) \\
\vdots & \ddots & \vdots \\
\phi(f_{\boldsymbol{\theta}_1}(\boldsymbol{x}_N), \mu_1, \sigma_1) & \ldots & \phi(f_{\boldsymbol{\theta}_M}(\boldsymbol{x}_N), \mu_M, \sigma_M) \\
\end{matrix}\right]
\]

In this work, the Gaussian basis function $\phi\left(\cdot\right)=\exp{\left(\frac{-1}{2\sigma_{j}^2}\left|f_{\boldsymbol{\theta}_j}(\boldsymbol{x_i})-\mu_{j}\right|^2\right)}$ have been used 
where the centers $c_{j}$ and scales $S_{j}$ parameters for the basis functions are computed from each column $j$ of the structured training dataset $\boldsymbol{D}(\varepsilon)$ 
and are computed by $\mu_{j}=\frac{1}{N}\sum_{i=1}^{N}{f_{\boldsymbol{\theta}_j}(\boldsymbol{x}_i)}$, and $\sigma_{j}=\sqrt{\sum_{i=1}^{N}\frac{\left(f_{\boldsymbol{\theta}_j}(\boldsymbol{x_i})-\mu_{j}\right)^2}{(N-1)}}$.

The s-RBFN formulation can now be expressed in matrix form as follows:

\begin{equation}
\label{matrixform}
    \boldsymbol{\hat{y}}=\boldsymbol{\Phi}\left(\boldsymbol{D(\varepsilon)}\right)\boldsymbol{w}=\left[\begin{matrix}\phi\left(f_{\boldsymbol{\theta}_1}\left(\boldsymbol{x}_1 \right),\mu_1, \sigma_1\right)&\ldots&\phi\left(f_{\boldsymbol{\theta}_M}\left(\boldsymbol{x}_1 \right), \mu_M, \sigma_M\right)\\\vdots&\ddots&\vdots\\\phi\left(f_{\boldsymbol{\theta}_1}\left(\boldsymbol{x}_N \right),\mu_1, \sigma_1\right)&\ldots&\phi\left(f_{\boldsymbol{\theta}_M}\left(\boldsymbol{x}_N \right), \mu_M, \sigma_M\right)\\\end{matrix}\right]\left[\begin{matrix}w_1\\\vdots\\w_M\\\end{matrix}\right]
\end{equation}

The optimal weights $\{w_i\}_{j=1}^M$ in (\ref{matrixform}) can now be simply obtained by least-squares with regularization parameter $\lambda_s$ for the structured model using:

\begin{equation}    \boldsymbol{w}=\left(\boldsymbol{\Phi}\left(\boldsymbol{D(\varepsilon)}\right)^{\rm T}\boldsymbol{\Phi}\left(\boldsymbol{D(\varepsilon)}\right)+\lambda_s\ast \boldsymbol{I}_{(mxm)}\right)^{-1}\boldsymbol{\Phi}\left(\boldsymbol{D(\varepsilon)}\right)^{\rm T}\boldsymbol{y}
    \label{weights}
\end{equation}

\begin{figure}[h!]
\setlength\abovecaptionskip{0\baselineskip}
	\centering
	\includegraphics[width=120mm]{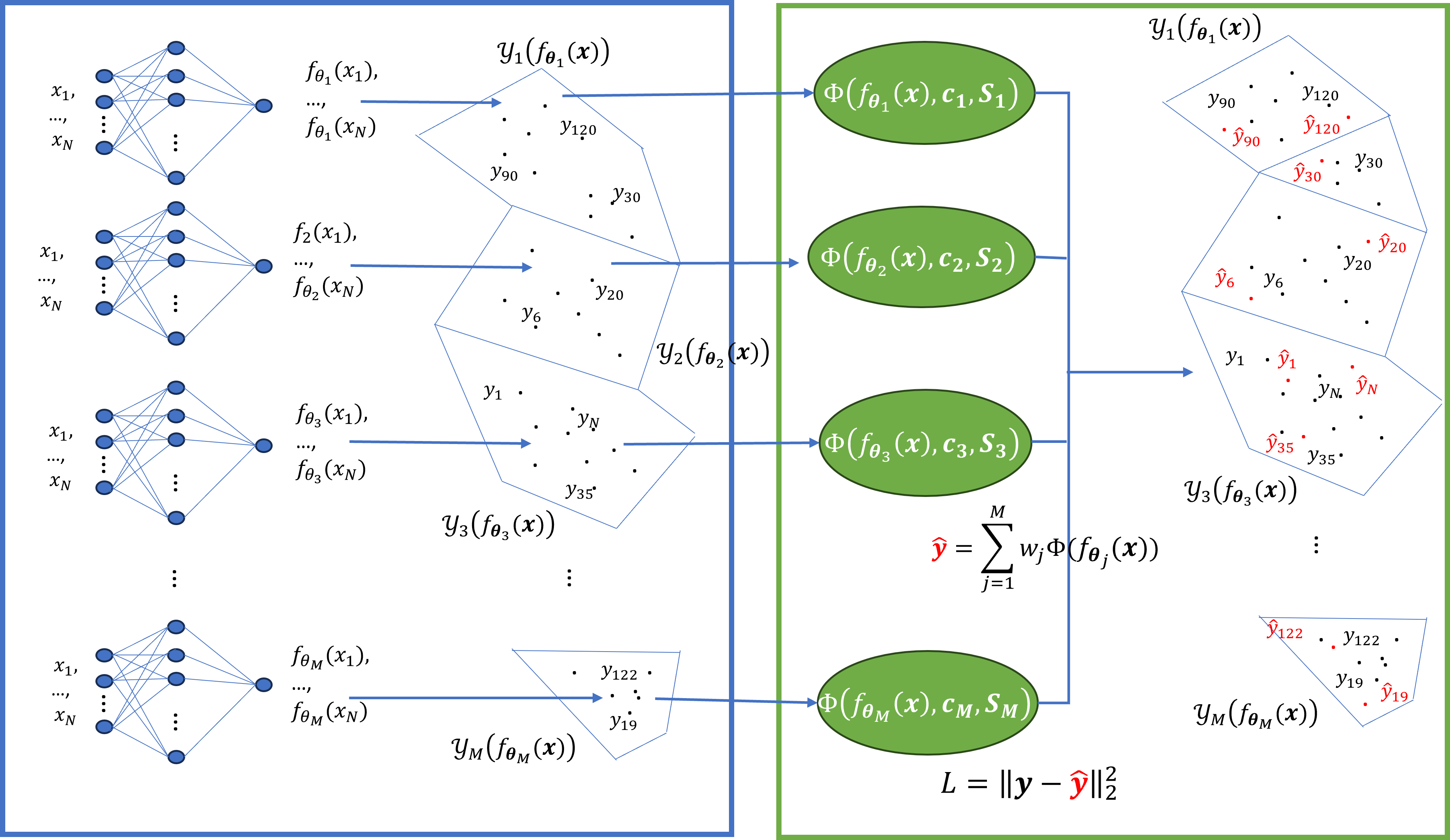}
	\caption{Model architecture with structured data obtained from neural networks' predictions and the ground-truth labels \(\boldsymbol{y}\) forming centroidal Voronoi tessellations based on the neural networks' losses (Left box). The s-RBFN uses these predictions to estimate the ground-truth labels \(\hat{\boldsymbol{y}}\), with \(L\) representing the s-RBFN norm loss (Right Box).
}
	\label{srbfnarchitecrure}
\end{figure}

The whole approach presented above has been summarized in the Figure \ref{srbfnarchitecrure} where the model is shown with the structured data obtained in the left box, using neural networks as predictors. The label data is assigned to a particular Voronoi tessellations depending on how far it is from the predictions of the base learners. This tessellation represents the multiple hypotheses prediction target values, \(\boldsymbol{y}\) (Left box in Figure \ref{srbfnarchitecrure}). Once training is completed, the predictions from the trained predictors, using all training instances as input, are used as input data for training a radial basis function network (s-RBFN) via least-squares. The estimates \(\boldsymbol{\widehat{y}}\) in Figure \ref{srbfnarchitecrure} of the multiple hypotheses prediction ground-truth labels \(\boldsymbol{y}\) are given by the output from the s-RBFN (Right box in Figure \ref{srbfnarchitecrure}). \(L\) represents the \(L_2\) norm loss between the ground-truth labels and their estimates.

\section{Experiments}
\label{secexp}

\subsection{Datasets}
An Air Quality dataset \cite{DEVITO2008750} and the Appliances Energy Prediction dataset \cite{CANDANEDO201781} are have been employed in this study. The first dataset consists of 9358 instances of hourly averaged responses from five metal oxide chemical sensors embedded in an air quality chemical multisensor device. The data was recorded from March 2004 to February 2005, and represents the longest freely available recordings of on-field responses from deployed air quality chemical sensor devices \cite{DEVITO2008750}. The goal is to predict absolute humidity values with the rest of variables in a multivariate regression problem. The second dataset consist of 10 minutes timestamps for 4.5 months making up over 20 thousand instances from 29 features. The goal is to predict energy appliances in a low energy building \cite{CANDANEDO201781}. \par

\subsection{Models Performance $\&$ Comparisons}
For the individual predictors, a 2-layer multi-layer perceptron (MLP) have been used with the number of neurons in each layer as $\kappa$, learning rates $\eta$, multiplicative factor of the initial weights $\chi$, and regularization parameters $\lambda_p$. For the s-RBFN model, the number of predictors or hypotheses is given by $M$, diversity parameter $\varepsilon$, and s-RBFN regularization parameters $\lambda_s$. All values used for the hyper-parameters are displayed in Table \ref{hyper}.

\begin{table}[h!]
\caption{Sets of values for the s-RBFN hyperparameters}
\label{Table1values}
\centering
\begin{subtable}[t]{0.9\textwidth}
\centering
\caption{$M$ number of hypotheses, $\kappa$ number of neurons per layer, $\eta$ learning rates for the predictors, $\chi$ is a multiplicative factor for random initial predictors' weights $\Theta$.}
\begin{tabular}{|c| c |c| c |}\hline 
$M$     & $\kappa$   &   $\eta$   &    $\chi$  \\
\hline  
$[2, 5, 10, 20, 35]$     & $[20, 200, 2000]$    &   $[0.03, 0.3]$    &    $[0.0001, 0.01, 0.1, 1]$  \\  \hline
\end{tabular}
\end{subtable}

\vspace{0.5cm}

\begin{subtable}[t]{0.9\textwidth}
\centering
\caption{$\varepsilon$ is the diversity parameter, $\lambda_p$ is the regularization parameter for the predictors, $\lambda_s$ is the regularization parameter for the s-RBFN.}
\begin{tabular}{|c| c |c| c |}\hline 
$\varepsilon$  &   $\lambda_p$  &  $\lambda_s$ \\
\hline  
$[0, 0.1, 0.35, 0.5]$   &   $[0, 0.0001, 0.01, 0.07]$   &  $[0, 3, 5]$ \\  \hline
\end{tabular}
\end{subtable}
\label{hyper}
\end{table}

For the experiments, the top performing models' versions from the original papers of the two used datasets \cite{DEVITO2008750,WB:2014,CANDANEDO201781}, are replicated for comparison (top competitors). These are the Linear Model (LM), Random Forest (RF), Gradient-Boost (Gboost), and Support Vector Machine Radial Basis Function (SVM-RBF). To elaborate, for the s-RBFN, the experiments are performed with 10 simulations for each combination of hyper-parameters from Table \ref{hyper}. The mean and standard deviations of the RMSE for each of the 10-folds are recorded as performance measures. In total, the experiments have been performed with 80 different model hyper-parameters' configurations (also including the single hypothesis $M=1$). Additionally, for further comparison, the bench-marking results using the baseline multiple hypothesis prediction (arithmetic combiner) model \cite{rupprecht2017learning} are also included, in which the ensemble of individual predictors forming Voronoi Tessellations as their arithmetic mean are employed.

\subsubsection{Absolute Humidity Prediction}
In Table \ref{tableauc_test}, the 10 cross-folds mean and standard deviation RMSE values for the top performing versions of all models on the test set are presented. The RMSE for the 80 different hyper-parameter configurations are computed and its first and third quartiles are shown in this table. For the rest of the models, 80 different hyper-parameters are applied for comparison.

The best model by generalization performance is the s-RBFN when the hyper-parameters are optimized. The SVM-RBF is the second best performing model. The arithmetic combiner has the lowest standard deviation and consequently has the smallest variation of the mean RMSE for all quartiles. The s-RBFN has a quarter of the 80 different hyper-parameter configurations' mean RMSE values lower than all other models except for the SVM-RBF, due to its higher standard deviation.

\begin{table}[]
\centering
\caption{Absolute humidity prediction: Mean and standard deviation of the 10-fold cross-validation RMSE for the models with the top-performing hyper-parameters configuration in generalization performance. First and third quartiles are shown for all models from 80 different hyper-parameter configurations.}
\begin{tabular}{|l|l|l|l|l|}
\hline
\textbf{Models}        & \textbf{Top Model}        & \textbf{std dev}          & \textbf{First Quartile}   & \textbf{Third Quartile}    \\
\hline
Linear Model          & 7692.78 & 1657.53 & 8189.84 & 10488.35 \\
\hline
SVM-RBF               & \textbf{29.83}   & \textbf{1.99}    & 34.80   & 37.65    \\
\hline
Random Forest         & 55.66   & 15.47   & 69.00   & 91.55    \\
\hline
Gradient Boosting     & 55.76   & 38.73   & 93.92   & 151.58   \\
\hline
Arithmetic Combiner   & 39.19   & \textbf{0.15}    & 41.75   & 43.93    \\
\hline
s-RBFN                & \textbf{22.46}   & \textbf{9.14}    & 38.98   & 54.71\\ 
\hline
\end{tabular}
\label{tableauc_test}
\end{table}

\begin{table}[ht]
 \centering
 \caption{Energy appliance prediction: Mean and standard deviation of the 10-fold cross-validation RMSE for the models with the top-performing hyper-parameters configuration in generalization performance. First and third quartiles are shown for all models from 80 different hyper-parameter configurations.}
\begin{tabular}{|l|l|l|l|l|}
\hline
\textbf{Models}        & \textbf{Top Model}       & \textbf{std dev}         & \textbf{First Quartile}  & \textbf{Third Quartile}  \\
\hline
Linear Model           & 281.76 & 297.69 & 321.39 & 803.31 \\
\hline
SVM-RBF                & \textbf{104.68} & \textbf{1.27}   & \textbf{107.26} & 109.01 \\
\hline
Random Forest          & 298.46 & 29.35  & 328.48 & 373.81 \\
\hline
Gradient Boosting      & 292.08 & 67.26  & 389.10 & 476.89 \\
\hline
Arithmetic Combiner    & 115.17 & \textbf{0.11}   & 128.54 & 144.83 \\
\hline
s-RBFN                 & \textbf{101.12} & \textbf{2.42}   & \textbf{102.36} & 109.96 \\
\hline
\end{tabular}
\label{tableenergy_test}
\end{table}

\subsubsection{Energy Appliance Prediction}

For the energy appliance dataset the same set of experiments are performed as for the air quality dataset. In line with the results displayed in Table \ref{tableauc_test}, in Table \ref{tableenergy_test} it can be seen how the s-RBFN is the best performing model with less standard deviation than in the previous dataset. This makes the model best performer in the first and third quartiles. The arithmetic combiner is the model with lowest standard deviation and the SVM-RBFN is the second best performing model, in line with the air absolute humidity prediction experiments.

Thus, both the dataset, it has been validated empirically that the s-RBFN is the best performing model in terms of generalization performance and for a range of different hyper-parameters.

\subsection{Diversity \& Generalization Performance}
\begin{figure}[h!]
\setlength\abovecaptionskip{0\baselineskip}
	\centering
	\includegraphics[width=90mm]{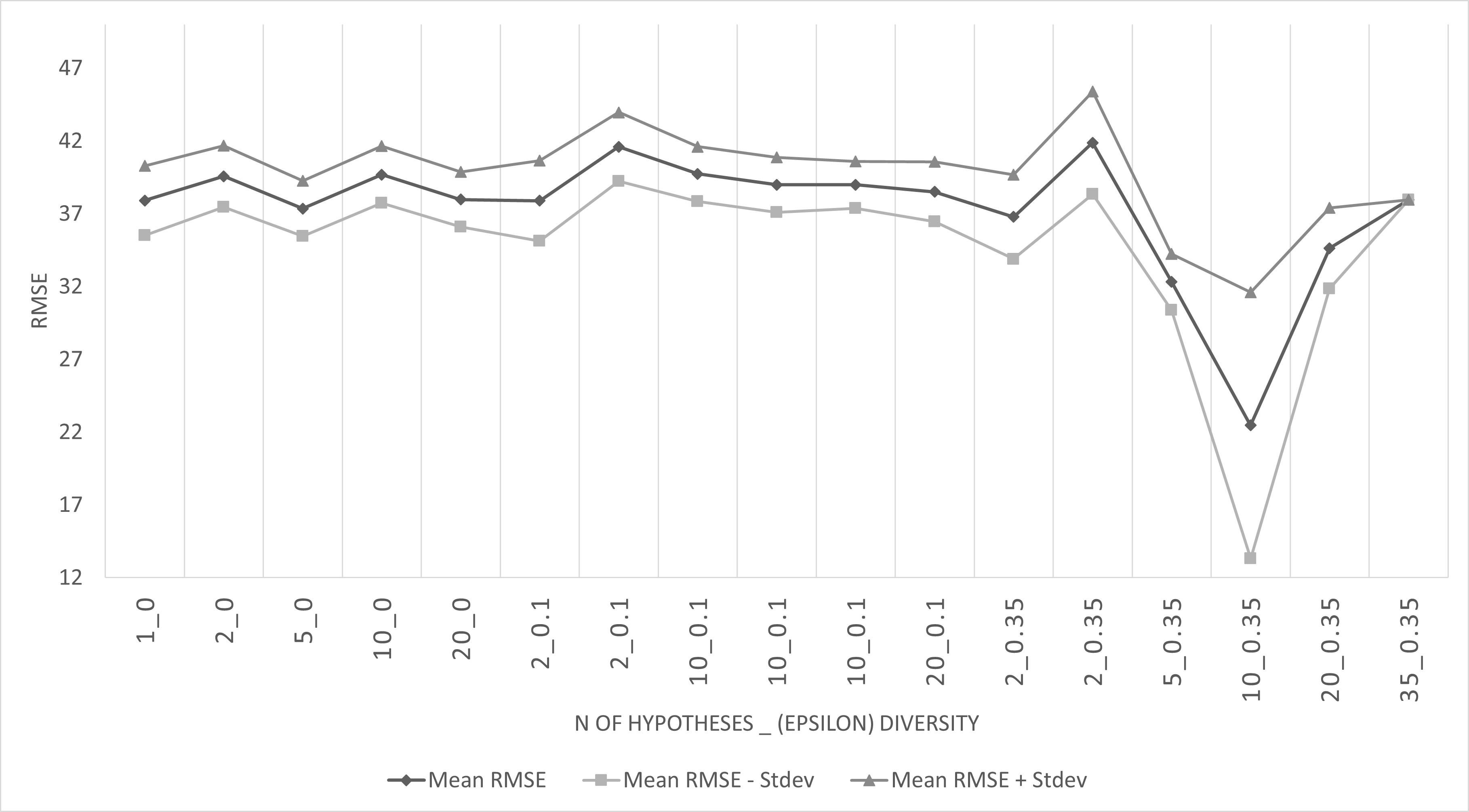}
 \vspace{2mm}
       \caption{Air quality test set: Mean RMSE and $90\%$ confidence interval from 10-fold cross-validation for different configurations for hyper-parameters $M$ and $\varepsilon$.}
	\label{DiversityAUC}
\end{figure}

\begin{figure}[h!]
\setlength\abovecaptionskip{0\baselineskip}
	\centering
	\includegraphics[width=90mm]{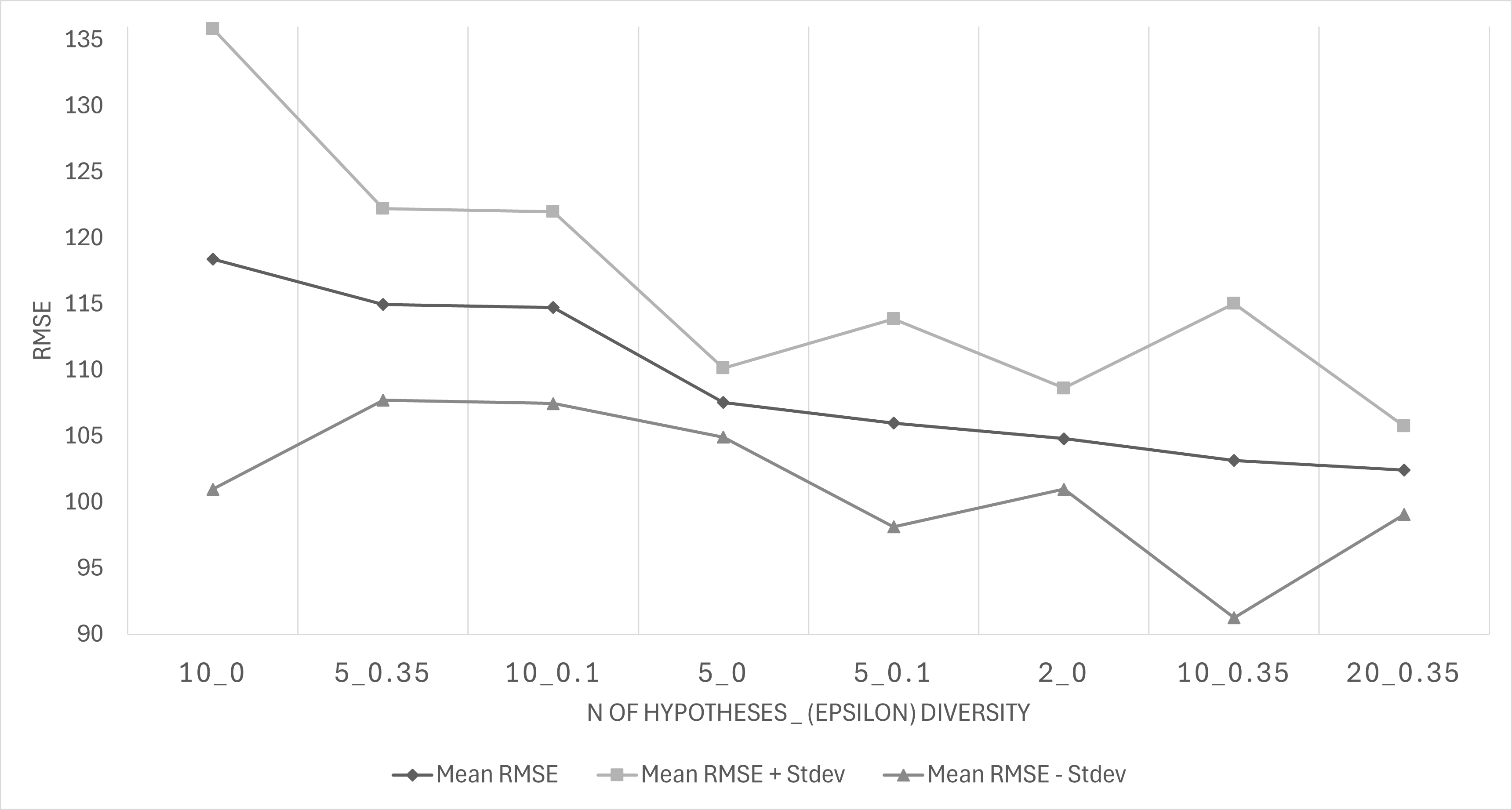}
 \vspace{2mm}
       \caption{Energy appliances test set: Mean RMSE and $90\%$ confidence interval from 10-fold cross-validation for different configurations for hyper-parameters $M$ and $\varepsilon$.}
	\label{DiversityEnergy}
\end{figure}

In this section, the hypothesis of the improvement in generalization performance of the s-RBFN for different values of the diversity parameter $\varepsilon$ and the number of hypotheses $M$ is verified. Figures \ref{DiversityAUC} and \ref{DiversityEnergy} show, for the air quality and energy appliances test sets respectively, the mean RMSE and $90\%$ confidence interval using 10-fold cross-validation for each hyper-parameter configuration, and for different values of the number of hypotheses $M$ and diversity parameter $\varepsilon$. The horizontal axis represents the pairs of hyper-parameters $M$ and $\varepsilon$.

The results in Figure \ref{DiversityAUC} indicate, for the absolute humidity prediction experiments with the air quality test set, that the generalization performance increases with the diversity parameter up to a certain number of hypotheses, but decreases if the number of hypotheses is too large. In this set of experiments the optimal pair for $M=10$ and $\varepsilon=0.35$ is well defined. For this pair of hyper-parameters the s-RBFN achieves the best performance, equal to the shown in Table \ref{tableauc_test}.  It can be shown that increasing $\varepsilon$ for two hypotheses worsen the generalization performance, meaning that a minimum number of hypotheses is needed for diversity to improve generalization capabilities.

In Figure \ref{DiversityEnergy}, the energy appliances prediction dataset shows the same conclusion with some different results. For relatively large number of hypotheses ($M=10, 20$) the s-RBFN achieves the best performance in generalization for relatively large $\varepsilon=0.35$. However, this improvement is not observed for $M=2$ and $M=5$, as for five hypotheses the best model has $\varepsilon=0.1$, with the case of $\varepsilon=0.35$ being worse than for the case of $\varepsilon=0$. It is reasonable to believe that for each number of hypotheses there is an optimal level of diversity, or $\varepsilon$, for the s-RBFN model. In the case of two hypotheses ($M=2$), there is no impact of diversity due to the low number of individual predictors. Moreover, for this case, the performance is very good, suggesting that while diversity can enhance generalization performance for a given number of hypotheses, there may be cases in which the individual predictor alone is good enough for prediction in the test set.


\subsection{Impact of Regularization}
\begin{figure}[h!]
    \centering
    \begin{minipage}{0.45\textwidth}
        \centering
        \includegraphics[width=\textwidth]{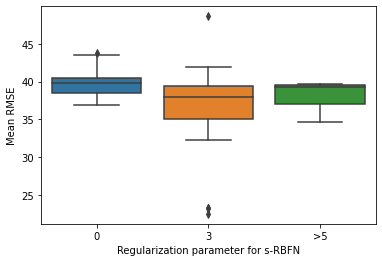}
        \subcaption{Mean RMSE}
        \label{fig:regair_mean}
    \end{minipage}\hfill
    \begin{minipage}{0.45\textwidth}
        \centering
        \includegraphics[width=\textwidth]{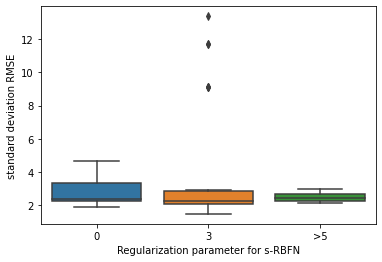}
        \subcaption{Standard deviation RMSE}
        \label{fig:regair_std}
    \end{minipage}
    \caption{Air quality test set: (a) Mean and (b) Standard deviation RMSE for 10-Fold cross-validation for different s-RBFN regularization parameters and hyper-parameter configurations.}
    \label{fig:regair}
\end{figure}

\begin{figure}[h!]
    \centering
    \begin{minipage}{0.45\textwidth}
        \centering
        \includegraphics[width=\textwidth]{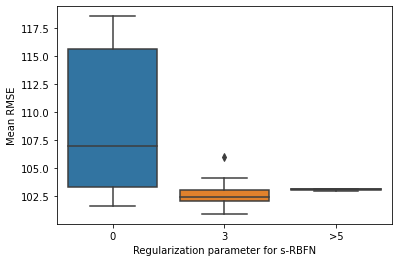}
        \subcaption{Mean RMSE}
        \label{fig:regenergy_mean}
    \end{minipage}\hfill
    \begin{minipage}{0.45\textwidth}
        \centering
        \includegraphics[width=\textwidth]{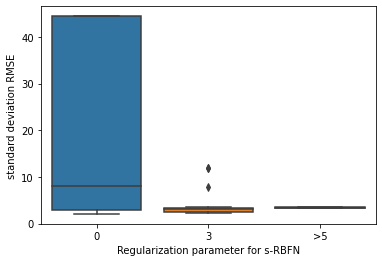}
        \subcaption{Standard deviation RMSE}
        \label{fig:regenergy_std}
    \end{minipage}
    \caption{Energy Appliance test set: (a) Mean and (b) Standard deviation RMSE for 10-Fold cross-validation for different s-RBFN regularization parameters and hyper-parameter configurations.}
    \label{fig:regenergy}
\end{figure}

In this section, the purpose is to understand the contribution of the regularization parameter for the s-RBFN in generalization performance. The regularization parameter \(\lambda_s\) has a clear effect in reducing the uncertainty of the hyper-parameters in the prediction of the structured ensemble model. For the air quality test set, in Figure \ref{fig:regair_mean}, it can be seen that for greater values of the regularization parameter, the mean RMSE for different hyper-parameter configurations remain more constant. Additionally, the standard deviation is lower for greater values of \(\lambda_s\), as shown in Figure \ref{fig:regair_std}. The same pattern is observed in the energy appliances test set with Figures \ref{fig:regenergy_mean} and \ref{fig:regenergy_std}. It can be concluded that the regularization parameter reduces the uncertainty of the s-RBFN hyper-parameters. It also improves the s-RBFN generalization performance, on average, for any value of the hyper-parameters.

In summary, the experiments demonstrate that diversity in structured ensemble models, particularly in the s-RBFN, is a distinctive feature of these architectures. There are instances where a single-hypothesis model may perform optimally. This indicates that diversity is not universally beneficial for enhancing generalization performance but rather improves performance contingent on a specific number of hypotheses. The experiments suggest there is indeed an optimal level of diversity, $\varepsilon$, for each number of hypotheses. Conversely, there exists a maximum number of hypotheses beyond which the performance of the ensemble model deteriorates, regardless of the $\varepsilon$ level, indicating limits to the benefits of diversification. Similarly, for $\varepsilon$ values exceeding $0.35$, there is a noticeable decline in overall generalization capabilities.
\section{Conclusion}
\label{conc}
This work introduces a novel structured ensemble model for single-output multiple hypotheses prediction. The presented model incorporates geometric properties of centroidal Voronoi tessellations with the individual predictors' losses during training. By altering the shape of the tessellations through a parametric mechanism, the diversity is introduced to the structured dataset for the s-RBFN model. It has been validated through experiments that the s-RBFN model surpasses other models in generalization performance across a range of hypotheses numbers and diversity parameters. This model is the fastest to train once the structured dataset is prepared using its closed-form expression. Additionally, it facilitates easy control over diversity in structured ensemble learning and multiple hypotheses prediction for single-output regression problems through the diversity parameter. It is crucial to analyze the appropriate number of hypotheses and diversity hyper-parameters for a specific dataset, as these are highly correlated with the generalization performance capabilities of the s-RBFN.

For future work, several areas can be explored to enhance structured ensemble models in multiple hypotheses prediction. For instance, this work uses tabular data for regression with a simple 2-layer network as individual predictors. It would be interesting to employ more datasets from other modalities, e.g., visual or text datasets and use deeper architectures. This would allow to further investigate the relationship between model diversity, complexity and the generalization performance.

\begin{credits}

\subsubsection{\discintname}
The authors have no competing interests to declare that are
relevant to the content of this article. 
\end{credits}
%
%
%
\bibliographystyle{splncs04}
\bibliography{mybibliography}
%





\end{document}